\pdfoutput=1

\documentclass[11pt]{article}

\usepackage{ACL2023}

\usepackage{times}
\usepackage{latexsym}
\usepackage{amsmath}
\usepackage{natbib}
\usepackage{multirow}
\usepackage{longtable, rotating, amssymb}
\usepackage{xcolor, color, soul}
\usepackage{enumitem}

\usepackage[T1]{fontenc}

\usepackage[utf8]{inputenc}

\usepackage{microtype}

\usepackage{inconsolata}

\usepackage{graphicx}
\usepackage{float}

\title{“When Words Fail, Emojis Prevail”: Generating Sarcastic Utterances with Emoji Using Valence Reversal and Semantic Incongruity}

\author{Faria Binte Kader$^*$, 
{\bf Nafisa Hossain Nujat$^*$,}
{\bf Tasmia Binte Sogir$^*$,} \\ 
{\bf Mohsinul Kabir,}
{\bf Hasan Mahmud,}
{\bf Kamrul Hasan}\\
Department of Computer Science and Engineering
\\Islamic University of Technology
\\Dhaka, Bangladesh\\
\texttt{\{faria, nafisa13, tasmia, mohsinulkabir, hasan, hasank\}@iut-dhaka.edu}\\
}
\begin{document}

\maketitle
\def\thefootnote{*}\footnotetext{These authors contributed equally to this work.}
\def\thefootnote{\arabic{footnote}}
\begin{abstract}
Sarcasm is a form of figurative language that serves as a humorous tool for mockery and ridicule. We present a novel architecture for sarcasm generation with emoji from a non-sarcastic input sentence in English. We divide the generation task into two sub tasks: one for generating textual sarcasm and another for collecting emojis associated with those sarcastic sentences. Two key elements of sarcasm are incorporated into the textual sarcasm generation task: valence reversal and semantic incongruity with context, where the context may involve shared commonsense or general knowledge between the speaker and their audience. The majority of existing sarcasm generation works have focused on this textual form. However, in the real world, when written texts fall short of effectively capturing the emotional cues of spoken and face-to-face communication, people often opt for emojis to accurately express their emotions. Due to the wide range of applications of emojis, incorporating appropriate emojis to generate textual sarcastic sentences helps advance sarcasm generation. We conclude our study by evaluating the generated sarcastic sentences using human judgement. All the codes and data used in this study has been made publicly available\footnote{\url{https://github.com/WrightlyRong/Sarcasm-Generation-with-Emoji}}. 
\end{abstract}

\section{Introduction}
Sarcasm is defined as the use of remarks that often mean the opposite of what is said in order to hurt someone's feelings or to criticize something in a humorous way\footnote{\url{https://dictionary.cambridge.org/}}. Sarcastic remarks are often challenging to interpret considering their literal meaning differs greatly from the speaker's actual intent. Compared to verbal or in-person conversations, textual sarcasm presents additional challenges due to the absence of visual cues, vocal tone etc.

\begin{table}[ht]
\centering
\small
\begin{tabular}{p{.35\linewidth}p{.5\linewidth}}
\hline
\textbf{Non-Sarcastic Input} & \textbf{Sarcastic Output with Emoji} \\
\hline
I really hate walking in the rain. & I really love the outdoors walking in the rain. I sat feeling thoroughly miserable.\includegraphics[scale=0.04]{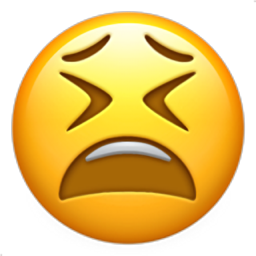} \\
\hline
Mom is in a bad mood today. & Happy mothers day mom is in a well mood today. She sounded tense and angry.\includegraphics[scale=0.04]{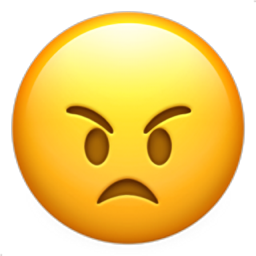} \\
\hline
That movie was bad. & That movie was awesome. Bad intelligence and political incompetence.\includegraphics[scale=0.04]{images/u1F620.png} \\
\hline
\end{tabular}
\caption{Sample sarcastic outputs with emoji generated from non-sarcastic inputs}
\label{sampleoutputs}
\end{table}

\noindent
The presence of sarcasm makes it significantly harder for machines to understand the actual meaning of the textual data. This has motivated research in detecting sarcasm in textual data. In order to train machines to detect sarcasm, we need quality datasets that represent different aspects of sarcasm in text. Even though we have an abundance of social media data and resources, it can be difficult to collect correctly labeled sarcastic texts. Instead, many research have tried to generate texts that can accurately express sarcastic notions \citep{joshi2015sarcasmbot, mishra2019modular, chakrabarty2020r}. Many studies have also investigated strategies in incorporating sarcasm generation into chatbots \citep{joshi2015sarcasmbot, joshi2017sarcasm}.\\
Emojis, small ideograms that represent objects, people, and scenes \citep{cappallo2015image2emoji}, are one of the key elements of a novel form of communication due to the advent of social media. Using emojis within texts can give us additional cues on sarcasm, replicating facial expressions and body language, etc. Incorporating emojis with texts for training will let the machines catch these cues easily \citep{bharti2016sarcastic}. \citet{subramanian2019exploiting} observed that when emojis were included in the sentence, their emoji-based sarcasm detection model performed noticeably better.\\
In this study, we propose a new framework in which when given a non-sarcastic text as input, the text is converted into a sarcastic one with emoji where the emoji will specifically help to identify the sarcastic intent of the text. Table \ref{sampleoutputs} shows a few sample non-sarcastic input and sarcastic output pairs with emoji. In order to implement the architecture, we have focused on two major components: Sarcastic text generation and Emoji prediction for the text. For textual sarcasm generation, we are incorporating the works of \citet{chakrabarty2020r} and \citet{mishra2019modular} and for Emoji prediction, a deep learning model fine tuned on OpenAI's CLIP (Contrastive Language-Image Pre-training)\footnote{\url{https://openai.com/research/clip}} \cite{radford2021learning} is used. The emoji prediction module along with the sarcasm generation module generates the final sarcastic text including emoji. This work provides two major contributions:
\begin{enumerate}
    \item Propose a novel multi-modular framework for sarcasm generation incorporating the reversal of valence and semantic incongruity characteristics of sarcasm while also including appropriate emojis.
    \item Create and publish a sarcastic corpora which can serve as valuable training data for sarcasm detection models.
\end{enumerate}
As far as our understanding goes, there has been no previous framework proposed on textual sarcasm generation that also incorporates emojis. This framework can aid downstream tasks by allowing a deeper understanding of sarcasm to produce more contextually relevant responses.

\begin{figure*}[ht]
    \centering
    \includegraphics[width=0.95\textwidth]{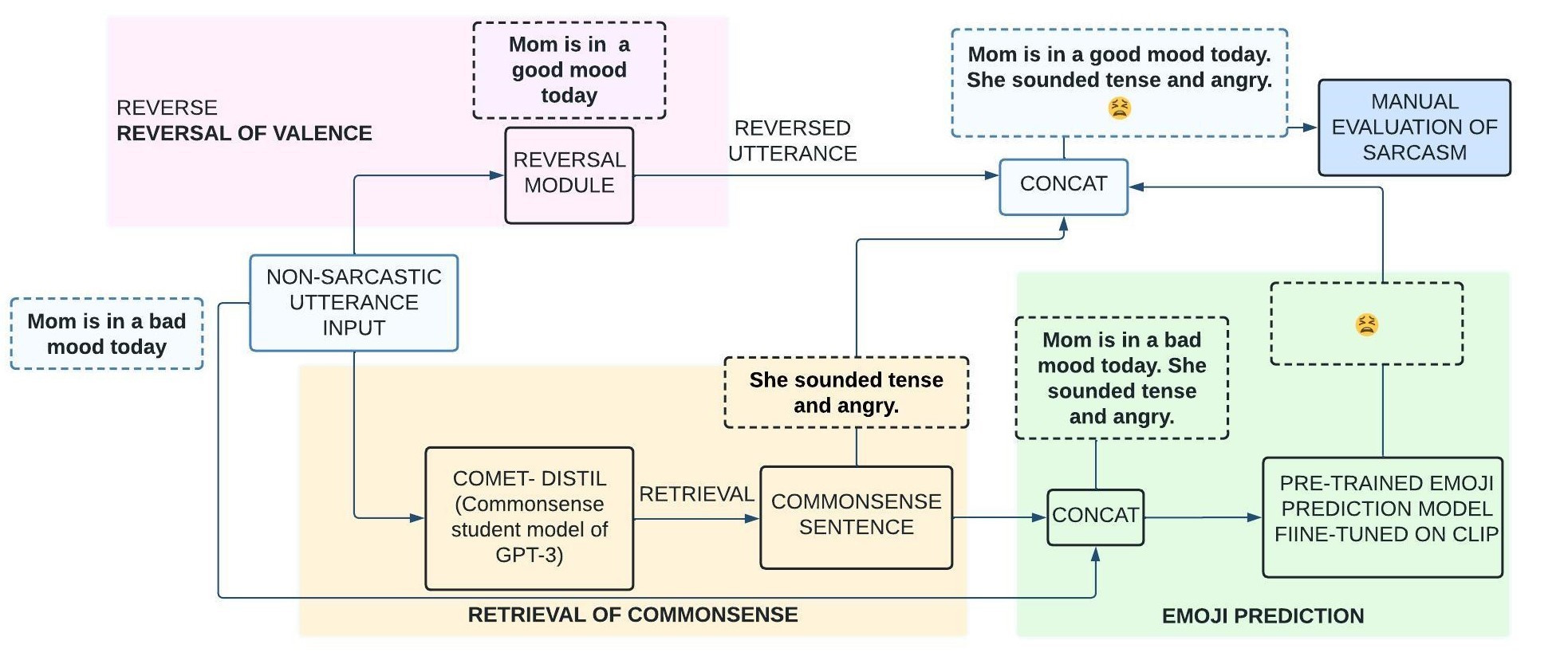}
    \caption{Model Architecture of the proposed system}
    \label{fig:architecture}
\end{figure*}

\section{Related Work}
Research on sarcasm have been a subject of interest for several decades. The following sub sections provide a brief overview of the past work done on different aspects of sarcasm.

\subsection{Studies on Sarcasm Detection}
Sarcasm detection is a classification task in its most typical form. From a given text, the task includes classifying the text as sarcastic or non-sarcastic. Sarcasm detection is a fairly recent but promising research field in the domain of Natural Language Processing. Nonetheless, it serves as a crucial part to sentiment analysis \citep{maynard2014cares}. \\
Most of these studies on sarcasm detection train and test on already available popular datasets such as the datasets used by \citet{riloff2013sarcasm}, \citet{khodak2017large} and \citet{cai2019multi}. We observed that Twitter is predominantly the most popular social media platform used for sarcasm detection datasets although Reddit, Amazon and a few discussion forums were also seen being used. We also saw a shift in Sarcasm detection methodologies from rule-based approaches \citep{riloff2013sarcasm, bharti2015parsing}, machine learning and deep learning approaches \citep{bharti2017sarcasm, poria2016deeper, ghosh2016fracking} to transformed based approaches \citep{dadu2020sarcasm, kumar2021adversarial}. We include two tables Table \ref{tabDetectDataset} and Table \ref{tabDetectMethod} summarizing the datasets and methodologies used in sarcasm detection in the appendix (Section \ref{sec:appendix}). \\
Recent works on sarcasm detection include frequent use of BERT \citep{savini2022intermediate, zhang2023stance, pandey2023bert}, multi-modal and cross-modal detection tasks \citep{liang2022multi, chauhan2022emoji, ding2022multimodal}, enhancement of sarcasm detection in complex expressions with sememe knowledge \citep{wen2022sememe}, study on the effect of foreign accent \citep{puhacheuskaya2022being}, use of vocal and facial cues \citep{aguert2022paraverbal} etc. Sarcasm and irony detection from languages other than English i.e. Chinese, Dutch, Spanish, Arabic, Romanian etc. have also been studied in recent works \citep{farha2020arabic, muaad2022artificial, maladry2022irony, wen2022sememe, ortega2022multi, buzea2022automatic}.

\subsection{Characteristics of Sarcasm}
Studies have identified a variety of potential sources for sarcasm. According to \citet{gerrig2000additive}, sarcasm stems from a situational disparity between what the speaker desires, believes, or expects and what actually happens. Incongruity between text and a contextual information is mentioned as a factor by \citet{wilson2006pragmatics}. Context Incongruity \citep{campbell2012there} is addressed in the works of \citet{riloff2013sarcasm} who suggests that sarcasm arises from a contrast between positive verbs and negative situation phrases. \citet{burgers2012verbal} formulates that for an utterance to be sarcastic, it needs to have one or more of these five characteristics:
\begin{enumerate}[noitemsep]
    \item the sentence has to be evaluative,
    \item it should be based on the reversal of valence of the literal and intended meanings,
    \item it should have a semantic incongruity with the context, which may consist of common sense or general information that the speaker and the addressee share,
    \item should be aimed at some target,
    \item  should be in some manner relevant to the communication scenario. Many studies focused on one or more of these characteristics.
\end{enumerate}

\subsection{Sarcasm Generation}
Compared to sarcasm detection, research on sarcasm generation is still in its early stages. \citet{joshi2015sarcasmbot} introduced SarcasmBot\footnote{\url{https://github.com/adityajo/sarcasmbot/}}, a chatbot that caters to user input with sarcastic responses. SarcasmBot is a sarcasm generation module with eight rule-based sarcasm generators where each of the generators produces a different type of sarcastic expression. During the execution phase, one of these generators is selected based on user input properties. Essentially, it yields sarcastic responses rather than converting a literal input text into a sarcastic one, the latter one being a common practice in future research. This method was later utilized in the author’s subsequent work \citep{joshi2017sarcasm} where they built SarcasmSuite, a web-based interface for sarcasm detection and generation.\\
The first work on automatic sarcasm generation conditioned from literal input was performed by \citet{mishra2019modular}. The authors relied on the Context Incongruity characteristic of sarcasm mentioned by \citet{riloff2013sarcasm} and employed information retrieval-based techniques and reinforced neural seq2seq learning to generate sarcasm. They used unlabeled non-sarcastic and sarcastic opinions to train their models, where sarcasm was formed as a result of a disparity between a situation's positive sentiment context and negative situational context. A thorough evaluation of the proposed system's performance against popular unsupervised statistical, neural, and style transfer techniques showed that it significantly outperformed the baselines taken into account.\\ 
\citet{chakrabarty2020r} introduced a new framework by incorporating context in the forms of shared commonsense or world knowledge to model semantic incongruity. They based their research on the factors addressed by \citet{burgers2012verbal}. Their architecture is structured into three modules: Reversal of Valence, Retrieval of Commonsense Context, and Ranking of Semantic Incongruity. With this framework they were able to simulate two fundamental features of sarcasm: reversal of valence and semantic incongruity with the context. However, they opted for a rule-based system to reverse the sentiments. The authors also noticed that in a few cases, the simple reversal of valence strategy was enough to generate sarcasm which meant the addition of context was redundant.\\
Recent similar works in the field include that of \citet{oprea2021chandler} where they developed a sarcastic response generator, Chandler, that also provides explanations as to why they are sarcastic. \citet{das2022unparalleled} manually extracted the features of a benchmark pop culture sarcasm corpus and built padding sequences from the vector representations’ matrices. They proposed a hybrid of four Parallel LSTM Networks, each with its own activation classifier which achieved 98.31\% accuracy among the test cases on open-source English literature. A new problem of cross-modal sarcasm generation (CMSG) that creates sarcastic descriptions of a given image was introduced by \citet{ruan2022describe}. However, these studies have only focused on generating textual sarcastic sentences, but as described by \citet{subramanian2019exploiting}, incorporating emojis improved the overall performance of sarcasm detection and thus can be a potential research scope.

\section{Methodology}
Our model architecture consists of 3 modules which are as follows: Reversal of Valence, Retrieval of Commonsense and Emoji Prediction. The Reversal of Valence module takes in a negative utterance and generates an utterance with positive sentiment. The Retrieval of Commonsense module outputs relevant commonsense context sentence which helps in creating a sarcastic situation. Lastly, the Emoji Prediction module generates an emoji which makes the overall output more sarcastic. With these three modules, we have incorporated two of the fundamental features of sarcasm: reversal of valence and semantic incongruity with the context. A diagram of the overall pipeline is demonstrated in Figure \ref{fig:architecture}. We describe the modules in details in the next few sub sections.

\subsection{Reversal of Valence} \label{Reversal of Valence}
In the work of \citet{chakrabarty2020r}, for the reversal of valence module, they have used a rule-based approach to manually reverse the sentiment of the negative sentence. But a rule-based model cannot reverse sentences that do not follow the traditional structure of sentences such as those used in social media. We have worked on this limitation of this current state-of-the-art sarcasm generation model where we replace their rule-based reversal module with a deep-learning reversal module inspired by the work of \citet{mishra2019modular}. This module is divided into two parts: Sentiment Neutralization and Positive Sentiment Induction.

\subsubsection{Sentiment Neutralization}
We implement the Sentiment Neutralization module to filter out the sentiment words from the input utterance, which results into a neutral sentence from a negative one. An example is shown in table \ref{neutralexample}.

\begin{table}[ht]
    \centering
    \small
    \begin{tabular}{p{0.2\textwidth} p{0.2\textwidth}}
    \hline
    \textbf{Negative Input} & \textbf{Neutral Output}\\
    \hline
    Is feeling absolutely bloated and fat from lack of a proper workout & Is feeling absolutely and from a proper workout\\
    \hline
    \end{tabular}
    \caption{Example of sentiment neutralization from input sentence}
    \label{neutralexample}
\end{table}

\begin{figure}[ht]
    \centering
    \includegraphics[width=80mm]{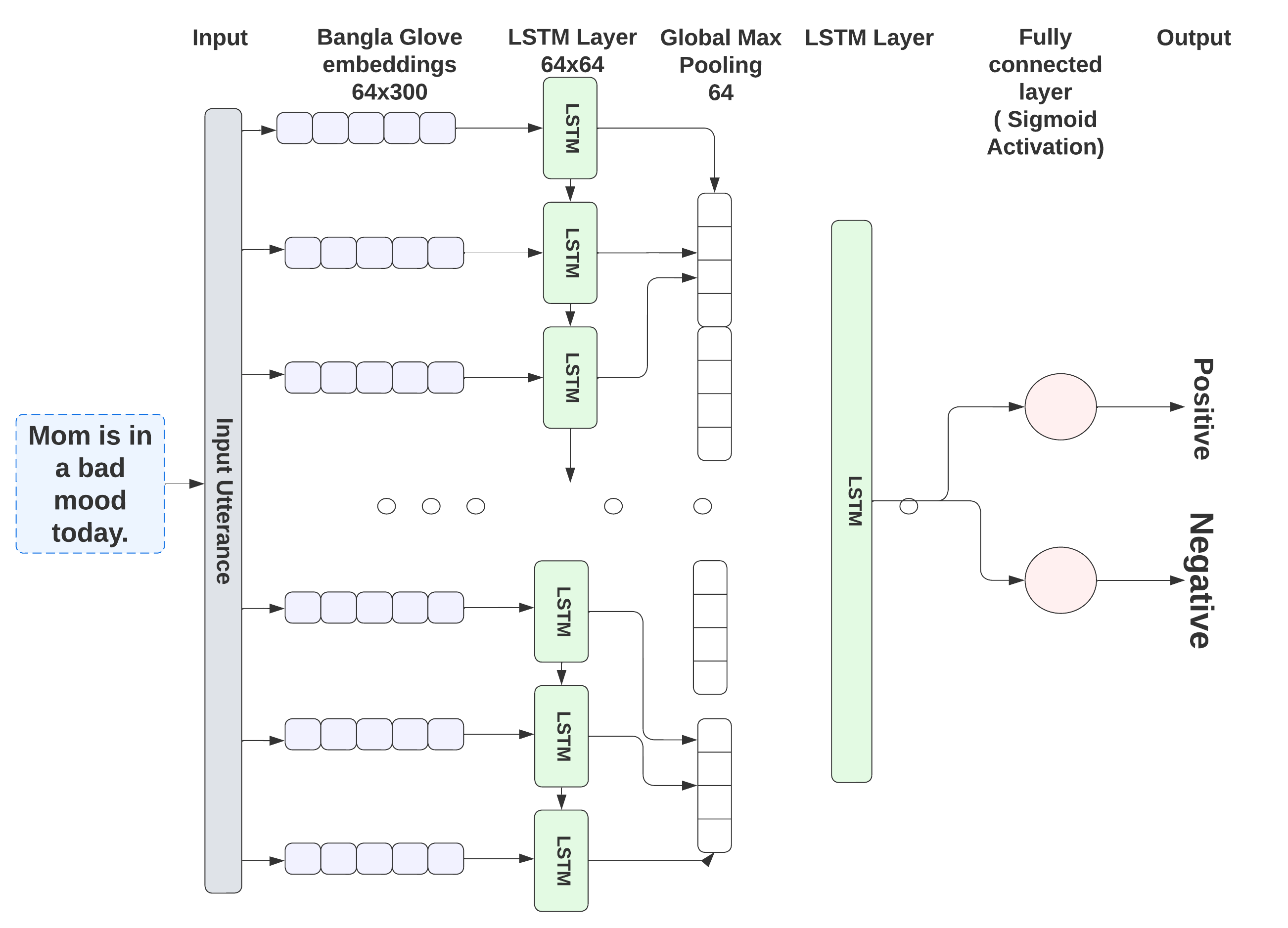}
    \caption{Sentiment detection model architecture for the Sentiment neutralization module}
    \label{fig:sentimentNeutralization}
\end{figure}

\noindent
The neutralization model is essentially a sentiment classification model which first detects the sentiment of the given utterance (positive/negative). This model consists of several LSTM layers and a self-attention layer. During testing, the self-attention vector is extracted as done by \citet{xu2018unpaired} which is then inversed and discretized as follows: 
\begin{equation}
    \hat{a_i}=
    \begin{cases}
      0, & \text{if $a_i > 0.95 * max(a)$}\\
      1, & \text{otherwise}
    \end{cases}
  \end{equation}
where $a_i$ is the attention weight for the $i^{th}$ word, and $max(a)$ gives the highest attention value from the current utterance. A word is filtered out if the discretized attention weight for that word is 0. The sentiment detection model architecture is shown in figure \ref{fig:sentimentNeutralization}.

\subsubsection{Positive Sentiment Induction}
The output from the Sentiment Neutralization module is fed to the Positive Induction module as input. The module takes in a neutral utterance and incorporates positive sentiment into the utterance and returns a sentence with positive sentiment. An example is shown in table \ref{positiveexample}. For this, we use Neural Machine Translation method built on OpenNMT framework \citep{klein2017opennmt} where we first train our model with a set of $<source, target>$ pairs where the source is a neutral sentence and target is its positive counter part. We use the Positive dataset provided by \citet{mishra2019modular} which includes a set of positive sentences. We pass this dataset through the sentiment neutralization module to get the neutral source sentence to its positive target sentence and use these $<source, target>$ pairs to train the positive induction module. The input sentences are transformed into embeddings that go through the translation encoders and decoders. The encoders and decoders are both built with LSTM layers. 
\begin{table}[ht]
    \centering
    \small
    \begin{tabular}{p{0.2\textwidth} p{0.2\textwidth}}
    \hline
    \textbf{Neutral Input} & \textbf{Positive Output}\\
    \hline
    Is feeling absolutely and from a proper workout & Is feeling absolutely amazing and high got away from a proper workout\\
    \hline
    \end{tabular}
    \caption{Example of positive sentiment induction from neutralized sentence}
    \label{positiveexample}
\end{table}


\subsection{Retrieval of Commonsense} \label{Retrieval of Commonsense}
\begin{figure}[ht]
    \centering
    \includegraphics[width=75mm,scale=1]{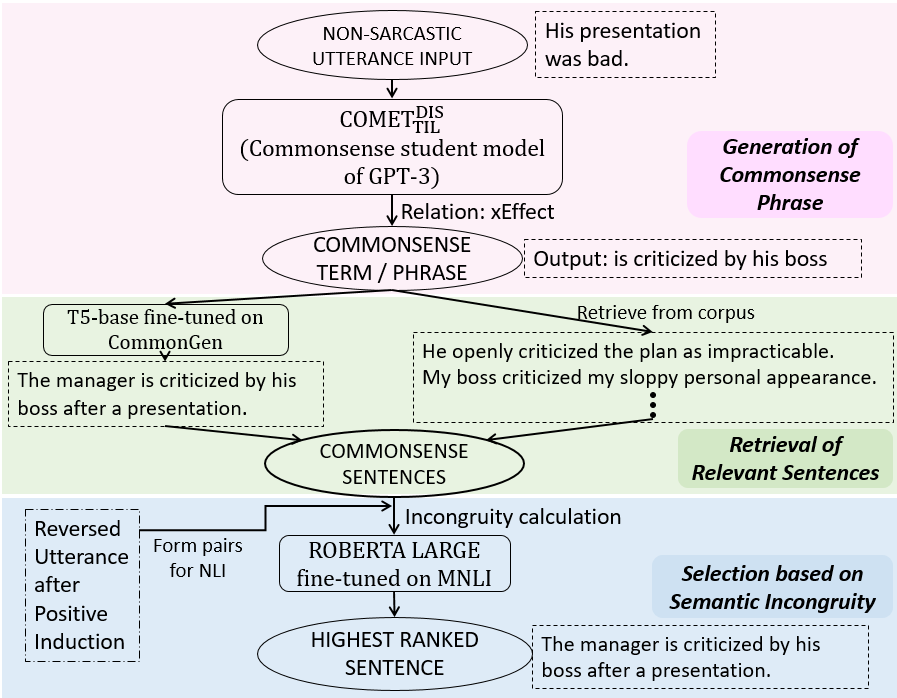}
    \caption{Model Architecture for Retrieval of Commonsense module}
    \label{fig:commonsense}
\end{figure}

This module is used to retrieve additional context for the sarcastic sentence based on commonsense knowledge.  
Figure \ref{fig:commonsense} demonstrates a schematic view of this module. We discuss the detailed process in the following sections. Additionally, we show an example input-output pair for this module in table \ref{commonsenseexample}. 
\begin{table}[ht]
    \centering
    \small
    \begin{tabular}{p{0.2\textwidth} p{0.2\textwidth}}
    \hline
    \textbf{Input} & \textbf{Commonsense Sentence}\\
    \hline
    His presentation was bad & The manager is criticized by his boss after a presentation\\
    \hline
    \end{tabular}
    \caption{Example of commonsense sentence generation from input sentence}
    \label{commonsenseexample}
\end{table}

\subsubsection{Generation of Commonsense Knowledge} \label{Generation of Commonsense Knowledge}
For generating commonsense knowledge context, COMET$_{\mathrm{TIL}}^{\mathrm{DIS}}$ \citep{west2021symbolic} is used. First, we feed the input sentence to COMET$_{\mathrm{TIL}}^{\mathrm{DIS}}$. COMET$_{\mathrm{TIL}}^{\mathrm{DIS}}$ is a machine trained 1.5B parameters commonsense model generated by applying knowledge distillation \cite{hinton2015distilling} on a general language model, GPT-3. It offers 23 commonsense relation types. For our study, we use the \textbf{xEffect} relation. From the three variants of COMET$_{\mathrm{TIL}}^{\mathrm{DIS}}$ (COMET$_{\mathrm{TIL}}^{\mathrm{DIS}}$, COMET$_{\mathrm{TIL}}^{\mathrm{DIS}}$ + critic$_{\mathrm{low}}$ and COMET$_{\mathrm{TIL}}^{\mathrm{DIS}}$ + critic$_{\mathrm{high}}$), we have chosen COMET$_{\mathrm{TIL}}^{\mathrm{DIS}}$ + critic$_{\mathrm{high}}$ for our work. The model returns a contextual phrase pertaining to the \textbf{xEffect} relation with the extracted words of the non-sarcastic sentence. For a non-sarcastic sentence “His presentation was bad”, COMET$_{\mathrm{TIL}}^{\mathrm{DIS}}$ predicts the contextual phrase with \textbf{xEffect} relation – ‘is criticized by his boss’.

\subsubsection{Retrieval of Relevant Sentences} \label{Retrieval of Relevant Sentences}
Once we have the inferred contextual phrase, we retrieve relevant sentences. For doing so, we imply 2 methods - 1. Retrieval from corpus and 2. Generation from the inferred phrase.
\begin{itemize}
    \item \textbf{Retrieval from corpus:} First, from the contextual phrase, we extract the keyword. Then using the keyword, we search for related sentences in a corpus. We use Sentencedict.com\footnote{\url{https://sentencedict.com/}} as the retrieval corpus. For filtering the retrieved sentences, two constraints are set - (a) the commonsense concept should appear at the beginning or at the end of the retrieved sentences; (b) to maintain consistency between the length of the non-sarcastic input and its sarcastic variant, sentence length should be less than twice the number of tokens in the non-sarcastic input. Next, we check the consistency of the pronoun in the retrieved sentence and the pronoun in the input sentence. If the pronoun does not match, we modify it to match the non-sarcastic text input. If the non-sarcastic input lacks a pronoun while the retrieved sentence does not, it is simply changed to “I”. These constraints for retrieving the sentences and the assessment of grammatical consistency are done following the work of \citet{chakrabarty2020r}. 
    \item \textbf{Generation from the inferred phrase:} Unlike the previous method, we keep the inferred phrase intact in this case. We first extract the \textit{Subject} of the non-sarcastic input. If the sentence contains no \textit{Subject}, we set it to 'I'. Then the auxiliary verb in the inferred context is checked and modified to match with that of the \textit{Subject}. Then we feed the \textit{Subject} and contextual phrase to a pre-trained sentence generation model\footnote{\url{https://huggingface.co/mrm8488/t5-base-finetuned-common_gen}}. The model fine-tunes Google's T5 on CommonGen \citep{lin2019commongen}. The model returns us a commonsense sentence based on the \textit{Subject} and contextual inference. For example - the \textit{Subject-inference} pair for the input “His presentation was bad” becomes [‘His’, ‘is criticized by his boss’], and from this collection of words, the sentence “The manager is criticized by his boss after a presentation.” is generated.
\end{itemize}

\subsubsection{Selection based on Semantic Incongruity}
The module in section \ref{Retrieval of Relevant Sentences} returns several sentences containing the context. Among them, we choose the sentence having the highest semantic incongruity with the sentence generated after the Reversal of Valence module. For calculating the semantic incongruity, following \citet{chakrabarty2020r}, we have used the RoBERTa-large \citep{liu2019roberta} model fine-tuned on the Multi-Genre NLI dataset \citep{williams2017broad}. Considering the non-sarcastic input “His presentation was bad”, the Retrieval of Relevant Sentences module yields a list of sentences such as - “The manager is criticized by his boss after a presentation”, “He openly criticized the plan as impracticable”, and “My boss criticized my sloppy personal appearance”. From these sentences, the highest ranked sentence, “The manager is criticized by his boss after a presentation”, is returned as the final output to this module as it contains the most semantic incongruity with the reversed sentence.

\subsection{Emoji Prediction}
\begin{figure}[ht]
    \centering
    \includegraphics[width=0.4\textwidth]{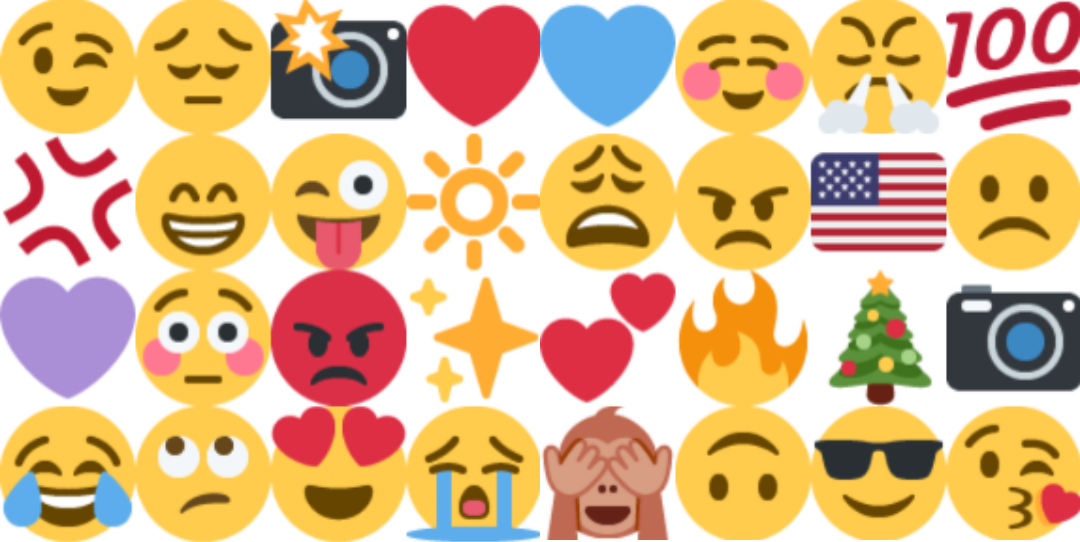}
    \caption{Set of 32 emojis}
    \label{fig:emojiset}
\end{figure}

In this module, we use a pre-trained emoji prediction model which is fine tuned on the CLIP (\citet{radford2021learning}) deep learning model by OpenAI to predict an emoji from a given input. After concatenating the non-sarcastic input and the context retrieved from the Retrieval of Commonsense module, we predict an emoji based on this concatenated sentence. The model employs a masked self-attention Transformer as a text encoder and a ViT-B/32 Transformer architecture as an image encoder. By using a contrastive loss, these encoders are trained to optimize the similarity of (image, text) pairs. One version of the implementation used a Vision Transformer and the other a ResNet image encoder. The variation with the Vision Transformer is used in this case. The dataset\footnote{\url{https://huggingface.co/datasets/vincentclaes/emoji-predictor}} used for fine-tuning the model consists of two columns: raw tweets and emoji labels. The emoji labels correspond to the appropriate one among a set of 32 emojis shown in figure \ref{fig:emojiset}.

\section{Experimental Setup}
The dataset, model configurations for the different modules, and the evaluation criteria for our work are all discussed in the following sub sections.

\subsection{Dataset}
For our experiments, we utilize the Positive and Negative sentiment corpora by \citet{mishra2019modular} which contains tweets and short snippets. Tweets have been normalized by eliminating hashtags, usernames, and conducting spell checking and lexical normalization using NLTK \cite{loper2002nltk}. After filtering out sentences longer than 30 words and running them through all three modules, we get the final dataset of 2k sarcastic sentences from the \citet{mishra2019modular} dataset. We have made our dataset\footnote{\url{https://github.com/WrightlyRong/Sarcasm-Generation-with-Emoji}} publicly available.

\begin{table*}[]
\centering
\tiny
\begin{tabular}{p{.15\linewidth}p{.1\linewidth}p{.3\linewidth}p{.055\linewidth}p{.045\linewidth}p{.045\linewidth}p{.075\linewidth}}
\hline
\textbf{Non-Sarcastic Utterance} & \textbf{System} & \textbf{Sarcastic Utterance} & \textbf{Sarcasticness} & \textbf{Creativity} & \textbf{Humor} & \textbf{Grammaticality} \\
\hline
\multirow{4}{0.85in}{Home with the flu.} & Full Model & Happy to be home with the fam. Being incarcerated-under the label of being mentally ill.\includegraphics[scale=0.04]{images/u1F62B.png} & 3.67 & 4.33 & 4 & 5 \\
\cline{2-7}
& Without Emoji & Happy to be home with the fam. Being incarcerated-under the label of being mentally ill. & 3.67 & 4.33 & 3.67 & 5 \\
\cline{2-7}
& Without Context & Happy to be home with the fam.\includegraphics[scale=0.04]{images/u1F62B.png} & 3.33 & 3 & 3 & 5 \\
\cline{2-7}
& R$^3$ \cite{chakrabarty2020r} & Home with the not flu. & 1.67 & 1.33 & 1.33 & 3 \\
\hline
\multirow{4}{0.85in}{The boss just came and took the mac away.} & Full Model & The boss just ended and took the mac away awesome. Angry is not the word for it - I was furious.\includegraphics[scale=0.04]{images/u1F620.png} & 5 & 5 & 4.67 & 4.33 \\
\cline{2-7}
& Without Emoji & The boss just ended and took the mac away awesome. Angry is not the word for it - I was furious. & 4 & 3.67 & 3 & 4.67 \\
\cline{2-7}
& Without Context & The boss just ended and took the mac away awesome.\includegraphics[scale=0.04]{images/u1F620.png} & 5 & 5 & 4.67 & 4.33 \\
\cline{2-7}
&R$^3$ \cite{chakrabarty2020r} &The boss just came and took the mac away. Angry is not the word for it - I was furious. & 1.67 & 2.33 & 1.67 & 5 \\
\hline
\multirow{4}{0.85in}{Friday nights are so boring when the boyfriend is working late and then i have to work at on saturday mornings.} & Full Model & Friday nights are so cute when the boyfriend is working rearrange and then i have to work at on mornings. At least they weren't bored. \includegraphics[scale=0.04]{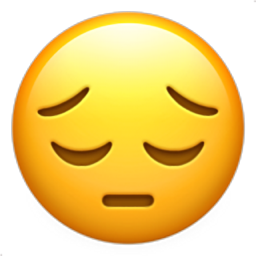} & 4 & 4 & 3.67 & 4 \\
\cline{2-7}
& Without Emoji & Friday nights are so cute when the boyfriend is working rearrange and then i have to work at on mornings. At least they weren't bored. & 4 & 4 & 3.67 & 4 \\
\cline{2-7}
& Without Context & Friday nights are so cute when the boyfriend is working rearrange and then i have to work at on mornings. \includegraphics[scale=0.04]{images/u1F614.png} & 4 & 4 & 3.67 & 4 \\
\cline{2-7}
&R$^3$ \cite{chakrabarty2020r} & Friday nights are so boring when the boyfriend is working early and then I have to work at on saturday mornings. Friday saw the latest addition to darlington's throbbing night life packed to the rafters. & 1.33 & 2 & 1.33 & 5 \\
\hline
\multirow{4}{0.85in}{Just finished workin bed feeling sick.} & Full Model & Just finished workin feeling good. My stomach heaved and I felt sick.\includegraphics[scale=0.04]{images/u1F62B.png} & 5 & 5 & 4.67 & 5 \\
\cline{2-7}
& Without Emoji & Just finished workin feeling good. My stomach heaved and I felt sick. & 5 & 5 & 4.67 & 5 \\
\cline{2-7}
& Without Context &Just finished workin feeling good.\includegraphics[scale=0.04]{images/u1F62B.png} & 3 & 3 & 3 & 5 \\
\cline{2-7}
&R$^3$ \cite{chakrabarty2020r} & Just finished workin bed feeling healthy. My stomach heaved and I felt sick. & 5 & 4.33 & 4.67 & 5 \\
\hline
\end{tabular}
\caption{Score comparison among the generated outputs from the different systems (Full model, Output without context, Output without emoji and the State-of-the-art model) on four categories}
\label{evaluationinputoutput}
\end{table*}

\subsection{Model Configurations}
The sentiment classification model of the neutralization module is trained on the sentiment dataset given by \citet{mishra2019modular} where the negative sentences are labeled as 1 and the positive sentences are labeled as 0. Each word in the input sentence is first encoded with one-hot encoding and turned into a K-dimensional embedding. Then, these embeddings go through an LSTM layer with 200 hidden units, a self-attention layer, an LSTM layer with 150 hidden units and finally a softmax layer. The classifier is trained for 10 epochs with a batch size of 32, and achieves a validation accuracy of 96\% and a test accuracy of 95.7\%.\\
The positive sentiment induction module is built on top of the OpenNMT 3.0 framework, and following \citet{mishra2019modular}, the embedding dimensions of the encoder and decoder is set to 500, with 2 LSTM layers each consisting of 500 hidden units. Training iteration is set to 100000 and early stopping is incorporated to prevent overfitting. After training, the model produced a corpus-BLEU score of 51.3\%.

\subsection{Evaluation Criteria}
For evaluating the performance of our proposed architecture we incorporate Human judgement. To assess the quality of the generated dataset we compare among 4 systems.
\begin{enumerate}
    \item \textbf{Full Model} contains all the proposed modules of the framework and generates the final dataset.
    \item \textbf{Without Emoji} system includes the context sentences along with the outputs from the reversal of valence module but does not contain any emoji that goes with each sarcastic sentence.
    \item \textbf{Without Context} system consists of generations from the reversal of valence module as well as emoji. It does not include any context.
    \item \textbf{R$^3$} is the state-of-the-art sarcasm generation system proposed by \citet{chakrabarty2020r}.
\end{enumerate}
To assess each of the four systems, we randomly choose 100 samples from our sarcastic dataset which totals to 400 output from the four systems. We evaluate these 400 generated sentences for comparing on the basis of the 4 above mentioned systems. \\ 
Following the evaluation approach proposed by \citet{chakrabarty2020r}, we evaluate the generated sentences on these criteria:
\begin{enumerate}
    \item Sarcasticness (“How sarcastic is the output?”),
    \item Creativity (“How creative is the output?”),
    \item Humour (“How funny is the output?”),
    \item Grammaticality (“How grammatically correct is the output?”).
\end{enumerate}
Previous studies on sarcasm generation have employed sarcasticness as a criterion for evaluating the effectiveness of the generated outputs \citep{mishra2019modular,chakrabarty2020r,das2022unparalleled}. As sarcasm exemplifies linguistic creativity \citep{gerrig1988beyond}, creativity has been proposed as a method for operationalizing the quality of sarcastic sentences by \citet{skalicky2018linguistic}. The association between humor and sarcasm is frequently mentioned in literature as well \citep{dress2008regional,lampert2006risky,leggitt2000emotional,bowes2011sarcasm}. The grammaticality criterion assesses the syntactic accuracy and conformity of the generated sentences.\\
Three human judges have been chosen to rate the outputs from the 4 systems on the 4 criteria mentioned. The label indicates a rating on a scale of 1 (not at all) to 5 (very). All 3 judges label each of the 400 sentences from the 4 systems. The human judges have been chosen based on their high efficiency in English, good grasp in understanding and differentiating between Creativity, Humor and Sarcasticness in English sentences. \\
To assess the inter-annotator agreement for the ratings, we incorporated the Intraclass Correlation Coefficient (ICC). ICC is a statistical measure used to assess the degree of agreement or correlation among the ratings given by different evaluators or raters for a certain category or metric. The agreement scores are shown in table \ref{icc}. The ICC score ranges between 0 and 1 where a higher score indicates a greater agreement among the raters. For all the four systems evaluated in our work, the ratings by 3 judges for the 4 evaluation criteria yield ICC scores above 0.9 in each case. A score above 0.9 indicates highly consistent observations and excellent agreement among the 3 judges.

\begin{table}[ht]
\centering
\small
\begin{tabular}{p{.10\textwidth}p{.065\textwidth}p{.065\textwidth}p{.065\textwidth}p{.065\textwidth}}
\hline
\textbf{System} & \multicolumn{4}{c}{\textbf{Intraclass Correlation Coefficient (ICC)}} \\
\cline{2-5}
\textbf{} & \textbf{S} & \textbf{C} & \textbf{H} & \textbf{G} \\
\hline
Full Model & 0.90 & 0.92 & 0.92 & 0.94 \\
\hline
Without Emoji & 0.95 & 0.96 & 0.95 & 0.92 \\
\hline
Without Context & 0.93 & 0.94 & 0.94 & 0.93 \\
\hline
R$^3$ \cite{chakrabarty2020r} & 0.97 & 0.97 & 0.97 & 0.97 \\
\hline
\end{tabular}
\caption{Intraclass Correlation Coefficient (ICC) scores on different metrics for the four systems. Here, S=Sarcasticness, C=Creativity, H=Humor, G=Grammaticality are the 4 evaluation criteria.}
\label{icc}
\end{table}

\noindent
Besides, human evaluation, we also evaluate our generated data against an emoji-based sarcasm detection model trained with existing emoji-based sarcastic dataset. For this, we utilize the work of \citet{subramanian2019exploiting} and use their proposed sarcasm detection model trained with their dataset. Their data samples were tweets with emojis scraped from Twitter and were labeled either 1 (sarcastic) or 0 (non-sarcastic). The model consists of a Bi-GRU with a text encoder and an emoji encoder. We add 2k non-sarcastic texts with our generated 2k sarcastic texts and test the model with these data. The model's performance is discussed in section \ref{result}.

\section{Experimental Results \& Analysis} \label{result}
\begin{table}[ht]
\centering
\small
\begin{tabular}{p{.15\textwidth}p{.05\textwidth}p{.05\textwidth}p{.05\textwidth}p{.05\textwidth}}
\hline
\textbf{System} & \multicolumn{4}{c}{\textbf{Variance$_{\mathrm{eval}}$}} \\
\cline{2-5}
\textbf{} & \textbf{S} & \textbf{C} & \textbf{H} & \textbf{G} \\
\hline
Full Model & 0.62 & 0.59 & 0.60 & 0.96 \\
\hline
Without Emoji & 0.74 & 0.73 & 0.65 & 0.96 \\
\hline
Without Context & 0.57 & 0.43 & 0.44 & 1.02 \\
\hline
R$^3$ \cite{chakrabarty2020r} & 1.48 & 1.17 & 1.16 & 0.99 \\
\hline
\end{tabular}
\caption{Variances among each evaluation criterion for each system. Here, S=Sarcasticness, C=Creativity, H=Humor,
G=Grammaticality are the 4 evaluation criteria.}
\label{variance}
\end{table}

\begin{table*}[]
\centering
\small
\begin{tabular}{p{.25\linewidth}p{.15\linewidth}p{.15\linewidth}p{.15\linewidth}p{.15\linewidth}}
\hline
\textbf{System} & \textbf{Sarcasticness} & \textbf{Creativity} & \textbf{Humor} & \textbf{Grammaticality} \\
\hline
Full Model & \textbf{3.44} & \textbf{3.29} & \textbf{3.16} & 3.72 \\
\hline
Without Emoji & 2.77 & 2.83 & 2.69 & 3.7 \\
\hline
Without Context & 3.1 & 2.99 & 2.88 & 3.72 \\
\hline
R$^3$ \cite{chakrabarty2020r} & 2.32 & 2.2 & 2.1 & \textbf{4.29} \\
\hline
\end{tabular}
\caption{Average ratings by human judges for outputs from the four systems}
\label{evaluation}
\end{table*}

\noindent
Table \ref{evaluationinputoutput} shows the comparison between a few sample sarcastic outputs across the various systems (our full model, output without the context, output without any emoji and lastly the state-of-the-art model \cite{chakrabarty2020r} on different measures (Sarcasticness, Creativity, Humor and Grammaticality). Each score is the average rating given by the three human judges. Table \ref{variance} shows the variances among each evaluation criterion for each of the four systems. The variances among the four criteria for the system R$^3$ are higher than all the other systems.\\
Table \ref{evaluation} shows the average ratings on 100 samples by human judges for generated sarcastic sentences from the four systems based on the four categories. Our full model achieves the highest average score among all the systems including the state-of-the-art sarcasm generation model by \citet{chakrabarty2020r} on three of the four categories except Grammaticality. Besides the full model, the without emoji system and without context system also outperform the state-of-the-art on Sarcasticness, Creativity and Humor. Our system lacks in Grammaticality due to the fact that we replace the rule based approach of the reversal of valence module by \citet{chakrabarty2020r} with a deep learning approach which results in a slightly more significant information loss. However, the rule based model performs worse in case of the other three categories as it fails to generalize on all types of sentence structures. It is apparent from the scores that context plays an important role in recognising a sarcastic sentence. Additionally, the notable improvement in the score for full model compared to the without emoji model suggests that emojis obviously help better detect the incongruity that exist in sarcastic utterances.\\
The emoji based sarcasm detection model by \citet{subramanian2019exploiting} gives an F1-score of 67.28\% and an ROC AUC score of 53.33\% on our generated data samples. It is to be noted that the model's training data samples have significantly different sentence structure than the test samples.

\section*{Conclusion}
We propose a novel multi-modular framework for sarcasm generation with emoji considering two key characteristics of sarcasm: reversal of valence and semantic incongruity between the sarcastic remark and the context. To generate sarcastic sentences, we first neutralize the input sentence's sentiment and then add positive sentiment to the sentence to reverse its meaning. We also incorporate a relevant emoji and its contextual information to enhance the sarcastic effect. We conclude by evaluating our model using human judgement.

\section*{Limitations}
Although our proposed architecture successfully generates emoji-based sarcastic sentences from non-sarcastic texts, in some cases, particularly longer sentences, adding commonsense context does not add much to make it more sarcastic as in such cases, the longer sentences already contain the contextual information. In future, we plan to modify our architecture in a way such that it can identify whether or not adding commonsense context would be necessary.\\
In our work, we have used COMET$_{\mathrm{TIL}}^{\mathrm{DIS}}$ to generate additional commonsense context. So the performance of our proposed architecture heavily depends on the accuracy of COMET$_{\mathrm{TIL}}^{\mathrm{DIS}}$. In future, we would like to find and incorporate better models for generating commonsense context.\\
The low grammaticality score by our final model is likely to be caused by the insufficient training data for the Positive Sentiment Induction module for which the model could not generalize properly. We believe that there is still room for improvement here by collecting and adding more training samples to improve the model's performance. To further fix the grammatical errors we plan to add another module after the Positive Induction module where the module will use a Transformer based grammar correction model which will take a sentence with bad grammar and output a grammatically correct sentence.\\
Lastly, our emoji prediction module only predicts one emoji per sentence. However, to make a sentence sarcastic, it is not uncommon to use more than one emoji. Hence, we plan to explore multi-label emoji prediction in the future.

\clearpage
\bibliography{anthology,custom}
\bibliographystyle{acl_natbib}

\clearpage
\onecolumn

\appendix

\section{Appendix}
\label{sec:appendix}

\begin{center}
\begin{longtable}
{|p{0.16\textwidth}|p{0.03\textwidth}|p{0.02\textwidth}|p{0.03\textwidth}|p{0.16\textwidth}|p{0.16\textwidth}|p{0.03\textwidth}|p{0.03\textwidth}|p{0.03\textwidth}|}
\caption{Summary of sarcasm detection datasets from different social media platforms}
\label{tabDetectDataset} \\
\hline
\textbf{}&\multicolumn{3}{|c|}{\textbf{Dataset}}&\textbf{}&\textbf{}&\multicolumn{3}{|c|}{\textbf{Annotation}}\\
\hline
\textbf{} 
& \begin{sideways}\textbf{Short Text}\end{sideways}
& \begin{sideways}\textbf{Long Text}\end{sideways}
& \begin{sideways}\textbf{Image}\end{sideways}
& \textbf{Samples}
& \textbf{Platform}
& \begin{sideways}\textbf{Manual}\end{sideways}
& \begin{sideways}\textbf{Hashtag}\end{sideways}
& \begin{sideways}\textbf{None}\end{sideways}
\\
\hline
\cite{filatova2012irony} & & \checkmark & & 1254 & Amazon & \checkmark & &\\
\hline
\cite{riloff2013sarcasm} & \checkmark & & & 1600 & Twitter & \checkmark & &\\
\hline
\cite{ptavcek2014sarcasm} & \checkmark & & & 920000 & Twitter & \checkmark & \checkmark &\\
\hline
\cite{barbieri2014modelling} & \checkmark & & & 60000 & Twitter & & \checkmark &\\
\hline
\cite{bamman2015contextualized} & \checkmark & & & 19534 & Twitter & & \checkmark &\\
\hline
\cite{amir2016modelling} & \checkmark & & & 11541 & Twitter & & \checkmark &\\
\hline
\cite{bharti2016sarcastic} & \checkmark & & & 1.5M & Twitter & & & \checkmark\\
\hline
\cite{joshi2016word} & \checkmark & & & 3629 & Goodreads & & \checkmark &\\
\hline
\cite{ghosh2016fracking} & \checkmark & & & 41000 & Twitter & & \checkmark &\\
\hline
\cite{poria2016deeper} & \checkmark & & & 100000 & Twitter & \checkmark & \checkmark &\\
\hline
\cite{schifanella2016detecting} & \checkmark & & \checkmark & 600925 & Instagram, Tumblr, Twitter & & \checkmark &\\
\hline
\cite{zhang2016tweet} & \checkmark & & & 9104 & Twitter & & \checkmark &\\
\hline
\cite{felbo2017using} & \checkmark & & & 1.6B & Twitter & & & \checkmark\\
\hline
\cite{ghosh2017magnets} & \checkmark & & & 41200 & Twitter & \checkmark & &\\
\hline
\cite{khodak2017large} & \checkmark & & & 533.3M & Reddit & \checkmark & &\\
\hline
\cite{oraby2017you} & & \checkmark & & 10270 & Debate forum & \checkmark & \checkmark &\\
\hline
\cite{prasad2017sentiment} & \checkmark & & & 2000 & Twitter & \checkmark & &\\
\hline
\cite{baziotis2018ntua} & \checkmark & & & 550M & Twitter & & & \checkmark\\
\hline
\cite{hazarika2018cascade} & \checkmark & & & 219368 & Reddit & \checkmark & &\\
\hline
\cite{ghosh2018sarcasm} & \checkmark & \checkmark & & 36391 & Twitter, Reddit, Discussion Forum & \checkmark & \checkmark &\\
\hline
\cite{ilic2018deep} & \checkmark & \checkmark & & 419822 & Twitter, Reddit, Debate Forum & \checkmark & \checkmark &\\
\hline
\cite{tay2018reasoning} & \checkmark & \checkmark & & 94238 & Twitter, Reddit, Debate Forum & \checkmark & \checkmark &\\
\hline
\cite{van2018semeval} & \checkmark & & & 4792 & Twitter & \checkmark & \checkmark &\\
\hline
\cite{wu2018thu_ngn} & \checkmark & & & 4618 & Twitter & \checkmark & \checkmark &\\
\hline
\cite{majumder2019sentiment} & \checkmark & & & 994 & Twitter & & \checkmark &\\
\hline
\cite{cai2019multi} & & & \checkmark & 24635 & Twitter & & \checkmark &\\
\hline
\cite{kumar2019sarcasm} & \checkmark & \checkmark & & 24635 & Twitter, Reddit, Debate Forum & & \checkmark &\\
\hline
\cite{subramanian2019exploiting} & \checkmark & \checkmark & & 12900 & Twitter, Facebook & & \checkmark &\\
\hline
\cite{jena2020c} & \checkmark & & & 13000 & Twitter, Reddit & \checkmark & \checkmark &\\
\hline
\cite{potamias2020transformer} & \checkmark & & & 533.3M & Twitter, Reddit & \checkmark & \checkmark &\\
\hline
\end{longtable}
\end{center}


\begin{center}
\begin{longtable}
{|p{0.12\textwidth}|p{0.09\textwidth}|p{0.19\textwidth}|p{0.09\textwidth}|p{0.08\textwidth}|p{0.08\textwidth}|p{0.08\textwidth}|}
\caption{Performance summary of various approaches used in sarcasm detection}
\label{tabDetectMethod} \\
\hline
&\textbf{Data}&\textbf{Architecture}&\multicolumn{4}{|c|}{\textbf{Performance}}\\
\hline
& & & \textbf{Accuracy} & \textbf{F1-Score} & \textbf{Precision} & \textbf{Recall} \\
\hline
\cite{davidov2010semi} & Tweets & SASI (Semi-supervised Algorithm for Sarcasm Identification) & 0.896 & 0.545 & 0.727 & 0.436\\
\hline
\cite{gupta2017crystalnest} & Tweets & CrystalNet & & 0.60 & 0.52 & 0.70\\
\hline
\cite{bharti2017sarcasm} & Tweets & PBLGA with SVM & & 0.67 & 0.67 & 0.68 \\
\hline
\cite{mukherjee2017detecting} & Tweets & Naive Bayes & 0.73 & & & \\
\hline
\cite{jain2017sarcasm} & Tweets & Weighted Ensemble & 0.853 & & 0.831 & 0.298 \\
\hline
\cite{poria2016deeper} & Tweets & CNN-SVM &  & 0.9771 &  & \\
\hline
\cite{ghosh2016fracking} & Tweets & CNN-LSTM-DNN &  & 0.901 & 0.894 & 0.912 \\
\hline
\cite{zhang2016tweet} & Tweets & GRNN & 0.9074  & 0.9074 &  &  \\
\hline
\cite{oraby2017you} & Tweets & SVM + W2V + LIWC &  & 0.83 & 0.80  & 0.86 \\
\hline
\cite{hazarika2018cascade} & Reddit posts & CASCADE & 0.79 & 0.86 & & \\
\hline
\cite{ren2018context} & Tweets & CANN-KEY &  & 0.6328 & &\\
\hline
& & CANN-ALL & & 0.6205 & & \\
\hline
\cite{tay2018reasoning} & Tweets, Reddit posts & MIARN & Twitter: 0.8647 & 0.86 & 0.8613 & 0.8579 \\
\hline
& & & Reddit: 0.6091 & 0.6922 & 0.6935 & 0.7005\\
\hline
\cite{ghosh2018sarcasm} & Reddit posts & multiple-LSTM & 0.7458 & 0.7607 & & 0.7762 \\
\hline
\cite{diao2020multi} & Internet arguments & MQA (Multi-dimension Question Answering model) & & 0.762 & 0.701 & 0.835 \\
\hline
\cite{kumar2020sarcasm} & Reddit posts & MHA-BiLSTM & & 0.7748 & 0.7263 & 0.8303 \\
\hline
\cite{kumar2019sarcasm} & Tweets & sAtt-BiLSTM convNet & 0.9371 & & & \\
\hline
\cite{majumder2019sentiment} & Text snippets & Multi task learning with fusion and shared attention & & 0.866 & 0.9101 & 0.9074\\
\hline
\cite{potamias2019robust} & reviews of laptops and restaurants & DESC (Deep Ensemble Soft Classifier) & 0.74 & 0.73 & 0.73 & 0.73 \\
\hline
\cite{srivastava2020novel} & Tweets, Reddit posts & BERT + BiLSTM + CNN & Twitter: 0.74 & & &\\
\hline
& & & Reddit: 0.639 & & &\\
\hline
\cite{gregory2020transformer} & Tweets, Reddit posts & Transformer ensemble (BERT, RoBERTa, XLNet, RoBERTa-large, and ALBERT) &  & 0.756& 0.758 & 0.767\\
\hline
\cite{potamias2020transformer} & Tweets, Reddit politics & RCNN-RoBERTa & Twitter: 0.91 & 0.90 & 0.90 & 0.90\\
\hline
& & & Reddit: 0.79 & 0.78 & 0.78 & 0.78\\
\hline
\cite{javdan2020applying} & Tweets & LCF-BERT & & 0.73 & & \\
\hline
& Reddit posts & BERT-base-cased &  & 0.734 & & \\
\hline
\cite{lee2020augmenting}  & Tweets, Reddit posts & BERT + BiLSTM + NeXtVLAD & Twitter & 0.8977 & 0.8747 & 0.9219\\
\hline
& & & Reddit & 0.7513 & 0.6938 & 0.8187\\
\hline
\cite{baruah2020context} & Tweets, Reddit posts & BERT-large-uncased & Twitter & 0.743 & 0.744 & 0.748\\
\hline
& & & Reddit & 0.658 & 0.658 & 0.658\\
\hline
\cite{avvaru2020detecting} & Tweets, Reddit posts & BERT & Twitter & 0.752 & &\\
\hline
& & & Reddit & 0.621 & & \\
\hline
\cite{jaiswal2020neural} & Tweets, Reddit posts & Ensemble of several combinations of RoBERTa-large & & 0.790 & 0.790 & 0.792\\
\hline
\cite{shmueli2020reactive} & Tweets & BERT & 0.703 & 0.699 & 0.70 0.7741 &\\
\hline
\cite{dadu2020sarcasm} & Tweets, Reddit posts & RoBERTa-large & Twitter & 0.772 & 0.772 & 0.772\\
\hline
& & & Reddit & 0.716 & 0.716 & 0.718\\
\hline
\cite{kalaivani2020sarcasm} & Tweets, Reddit posts & BERT & Twitter & 0.722 & 0.722 & 0.722\\
\hline
& & & Reddit & 0.679 & 0.679 & 0.679\\
\hline
\cite{naseem2020towards} & Tweets & T-DICE + BiLSTM + ALBERT & 0.93 & 0.93 & &\\
\hline
\cite{dong2020transformer} & Tweets, Reddit posts & context-aware RoBERTa-large & Twitter & 0.783 & 0.784 & 0.789\\
\hline
& & & Reddit & 0.744 & 0.745 & 0.749\\
\hline
\cite{kumar2020transformers} & Tweets, Reddit posts & context-aware RoBERTa-large & Twitter & 0.772 & 0.773 & 0.774\\
\hline
& & & Reddit & 0.691 & 0.693 & 0.699\\
\hline
\cite{kumar2021adversarial} & Tweets & AAFAB (Adversarial and Auxiliary Features-Aware BERT) & & 0.7997 & 0.8101 & 0.7896\\
\hline
\cite{lou2021affective} & Tweets, Reddit posts & ADGCN-BERT (Affective Dependency Graph Convolutional Network) & Twitter: 0.9031 & 0.8954 & & \\
\hline
& & & Reddit: 0.8077 & 0.8077 & & \\
\hline
\end{longtable}
\end{center}

\end{document}